# A Dual-Graph Spatiotemporal GNN Surrogate for Nonlinear Response Prediction of Reinforced Concrete Beams under Four-Point Bending

Zhaoyang Ren [1], Qilin Li [1, *]

[1] Discipline of Computing, School of Electrical Engineering, Computing and Mathematical Sciences, Curtin University, Australia

[*] Corresponding author: Qilin.Li@curtin.edu.au

## Abstract

High-fidelity nonlinear finite-element (FE) simulations of reinforced-concrete (RC) structures are still costly, especially in parametric settings where loading positions vary. We develop a dual-graph spatiotemporal GNN surrogate to approximate the time histories of RC beams under four-point bending. To generate training data, we run a parametric Abaqus campaign that independently shifts the two loading blocks on a mesh-aligned grid and exports full-field responses at fixed normalized loading levels. The model rolls out autoregressively and jointly predicts nodal displacements, element-wise von Mises stress, element-wise equivalent plastic strain (PEEQ), and the global vertical reaction force in a single multi-task setup. A key motivation is the peak loss introduced when element quantities are forced through node-based representations. We therefore couple node- and element-level dynamics using two recurrent graph branches: a node-level graph convolutional gated recurrent unit (GConvGRU) for kinematics and an element-level GConvGRU for history-dependent internal variables, with global force predicted through pooling on the element branch. In controlled ablations, removing the Element to Node to Element pathway improves peak-sensitive prediction in localized high-gradient stress/PEEQ regions without degrading global load–displacement trends. After training, the surrogate produces full trajectories at a fraction of the cost of nonlinear FE, enabling faster parametric evaluation and design exploration.

## 1. Introduction

Accurately predicting the nonlinear response and damage evolution of reinforced concrete (RC) structures remains challenging. Classical analytical solutions can be effective for simple geometries and linear elastic materials, but they are not sufficient for quasi-brittle RC, where stiffness degradation, cracking, and localized failure govern the response. Experiments provide essential evidence of load–displacement behaviour and failure modes, yet it is difficult to obtain continuous full-field internal variables

(e.g., stress and plastic strain) throughout fracture. High-fidelity finite element (FE) analysis has therefore become a standard tool for simulating crack initiation, plasticity development, and global structural response [1-5]. However, nonlinear FE analyses often require fine meshes and incremental–iterative solution procedures (e.g., Newton–Raphson) to capture localization and history dependence [6, 7]. This computational cost limits their use in rapid parametric studies, uncertainty quantification, and large-scale design exploration [8, 9].

To reduce reliance on repeated FE simulations, data-driven surrogate models have been explored for predicting structural capacity, dynamic response, and failure mechanisms [10-13]. Conventional deep learning models can accelerate inference after training [14-16], but their reliance on fixed-length vectors or regular grids makes it difficult to accommodate irregular FE meshes and varying discretisation. Related approaches such as PINNs and operator learning introduce physical constraints [17-20], but they can still struggle to generalize across mesh irregularity and strongly nonlinear, path-dependent responses [21, 22].

Graph neural networks (GNNs) address this representation issue by learning directly on mesh-based graphs, where nodes or elements form vertices and physical adjacency defines edges. Message passing provides a mechanism to encode localized interactions consistent with how information propagates in numerical simulations [23, 24]. In computational mechanics, GNN-based models have been applied to multiscale stress prediction, hyper-elasticity, plasticity, and spatiotemporal dynamics across different systems [25-29], and can perform autoregressive rollouts to recover full-field trajectories at reduced cost [25, 26].

Despite recent progress, most mesh-based GNN surrogates still rely on a single-scale, node-only representation [26, 27, 29]. This choice is convenient for kinematic fields such as displacement and velocity, but it is less suitable for history-dependent internal variables. In standard FEM, von Mises stress and equivalent plastic strain (PEEQ) are evaluated at element integration points rather than at nodes [30]. Node-only surrogates therefore require an Element to Node projection and a subsequent Node to Element interpolation for supervision. Such Element to Node to Element averaging is known to smooth sharp spatial gradients and suppress peaks [31, 32]. For RC, where crushing and yielding are highly localized, this peak loss can bias the predicted evolution of stress and plasticity and reduce surrogate fidelity.

To avoid this projection bottleneck, we propose a dual-graph spatiotemporal surrogate based on a Dual-Graph GConvGRU. The model maintains two coupled representations of the FE mesh: a node graph for the temporal evolution of kinematic quantities (e.g., nodal displacement) and an element graph for history-dependent internal variables (stress and PEEQ). The element branch also supports regression of the global vertical reaction force through pooling. We train the surrogate in a unified multi-task learning (MTL) setting [33], jointly supervising nodal, element-wise, and global outputs to exploit cross-scale correlations.

The main contributions are summarized as follows:

- A coupled node–element recurrent GNN is developed to avoid Element to Node to Element projection; controlled ablations quantify the associated peak attenuation (about 20%) and show improved peak-sensitive prediction with an RMSE reduction of approximately 29.5% relative to a single-graph baseline.
- A single spatiotemporal multi-task objective is formulated to jointly supervise nodal displacement, element-wise stress/PEEQ, and global reaction forces, allowing shared representations across kinematic, internal-variable, and global responses.
- A high-fidelity dataset of 190 parametrically varied Abaqus simulations of RC beams under four-point bending is constructed to support stable spatiotemporal learning and reproducible evaluation.
- After training, full response trajectories can be inferred with an approximately two-order-of-magnitude speedup over nonlinear FE analyses, enabling rapid parametric evaluation and design iteration.

The paper is structured as follows: Section 2 details the parametric FE dataset generation, followed by the formulation of our dual-graph surrogate and multi-task training strategy in Section 3. Section 4 presents predictive performance and ablation studies, and Section 5 concludes the work.

## 2. Benchmark and FE Dataset Generation

### 2.1 Reference experiment (CL30)

This study adopts a published four-point bending experiment as the physical benchmark to ensure that the finite element (FE) dataset used for GNN training is grounded in a realistic structural response. The reference experiment was reported by Wang et al. [34], which investigated reinforced concrete (RC) beams (including CFRP-strengthened cases) under sustained and subsequent loading. In the present work, the unstrengthened control beam CL30 is selected as the baseline specimen and reproduced in Abaqus to generate simulation data for surrogate learning. The CFRP-strengthening details and load-history scenarios in the original study are not considered here; instead, CL30 serves as a validated prototype for mesh-based learning.

The CL30 beam is a prismatic RC member with overall dimensions of 150 mm × 250 mm × 2700 mm (width × depth × length) and a clear span of 2400 mm under a simply supported configuration. Longitudinal reinforcement comprises three $\phi 14$ deformed bars in tension and two $\phi 8$ bars in compression, while transverse reinforcement consists of $\phi 6$ stirrups at 200 mm spacing, as shown in Figure 2.1. The concrete strength grade is C30, and the geometry and reinforcement layout follow the specimen drawing in Wang et al. [34], which is used consistently as the baseline for FE reproduction.

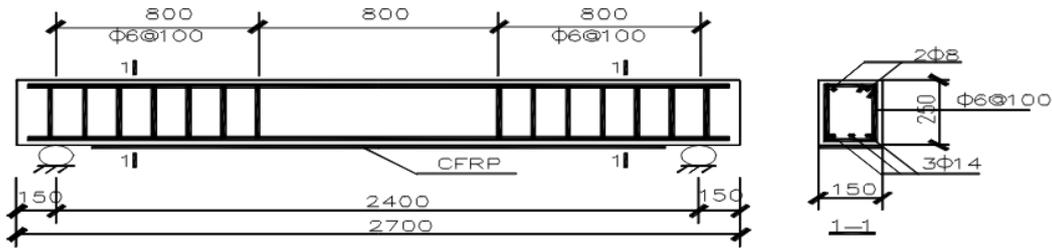

Figure 2.1 Geometry and reinforcement details of the CL30 reference beam.

Material properties for the reference specimen include the concrete compressive strength and elastic modulus, as well as the mechanical properties of reinforcing steel. The concrete compressive strength is 40.3 MPa with an elastic modulus of 32.7 GPa [35]. For reinforcing steel, the $\phi 14$ bars have a yield strength of 365.9 MPa and an ultimate strength of 535.9 MPa with an elastic modulus of 200 GPa; the $\phi 8$ bars have a yield strength of 352.1 MPa and an ultimate strength of 523.9 MPa with an elastic modulus of 210 GPa. These reported values are used as primary references when defining the FE material models in Abaqus, as shown in Table 1.

Table 1 Material properties of concrete and reinforcing steel

| **Material** | $f_y$(MPa) | $f_y$(MPa) | $\varepsilon(10^{-6})$ | $E$(GPa) |
| --- | --- | --- | --- | --- |
| Concrete |  | 40.3 |  | 32.7 |
| Steel bar ($\phi 14$) | 365.9 | 535.9 | 1829.5 | 200 |
| Steel bar ($\phi 8$) | 352.1 | 523.9 | 1676.7 | 210 |

The reference test used a simply supported four-point bending setup, with two symmetrically applied concentrated loads creating a constant-moment region between the loading points. Measured responses included the load–displacement curve, reinforcement strain, and crack development. For the CL30 specimen, the reported yield load is 85 kN at a midspan displacement of 10.02 mm, and the ultimate load is 102 kN at a displacement of 33.4 mm. The observed failure mode is concrete crushing.

These measurements provide a well-documented benchmark for constructing and validating the baseline Abaqus model. Based on the validated baseline, a parametric FE dataset is subsequently generated by varying the load locations, and a GNN surrogate is trained to predict full-field displacements and stress-related quantities across a range of loading configurations.

## 2.2 Abaqus finite element model of the baseline RC beam

This section describes the Abaqus finite element model used to reproduce the baseline RC beam response under symmetric four-point bending. The model provides the high-fidelity data that are later exported and used to train the GNN surrogate.

2.2.1 Model parts and assembly

The FE model consists of a concrete solid part, an embedded reinforcement part, two loading blocks, and two supports. The loading blocks are positioned symmetrically on the top surface to represent the two-point loading configuration. The supports are placed to achieve a clear span of 2400 mm. The assembled model is shown in Figure 2.2.

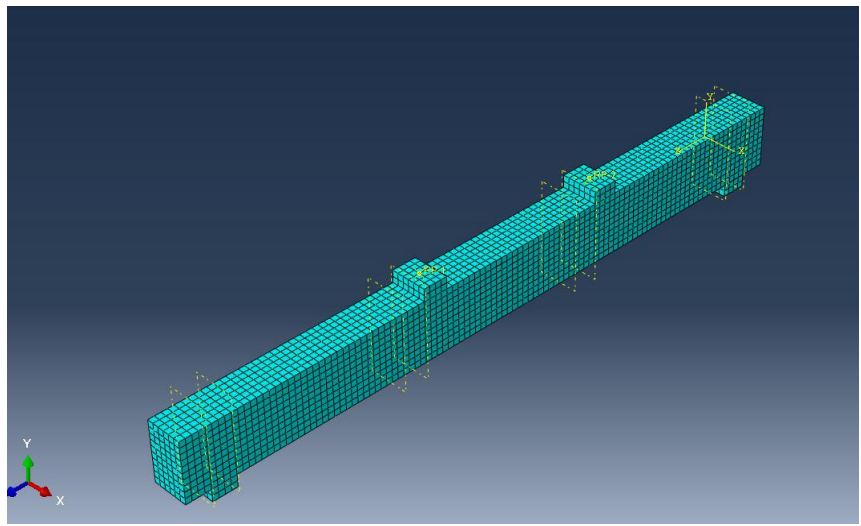

Figure 2.2 Assembled Finite Element Model of the RC Beam.

2.2.2 Element types and mesh

Concrete is discretised using eight node reduced integration brick elements C3D8R. Steel reinforcement, including longitudinal bars and stirrups, is discretised using two node three-dimensional truss elements T3D2. A uniform global mesh size of 25 mm is adopted for the concrete. This mesh size is used for all baseline simulations to balance computational cost and the ability to capture the nonlinear flexural response.

### 2.2.3 Reinforcement embedding

Reinforcement is embedded into the concrete using the Embedded Region constraint. This setting assumes perfect bond between steel and concrete, Wireframe as show in Figure 2.3. No bond slip is modelled. The embedded constraint ensures that reinforcement deformation is compatible with the host concrete elements.

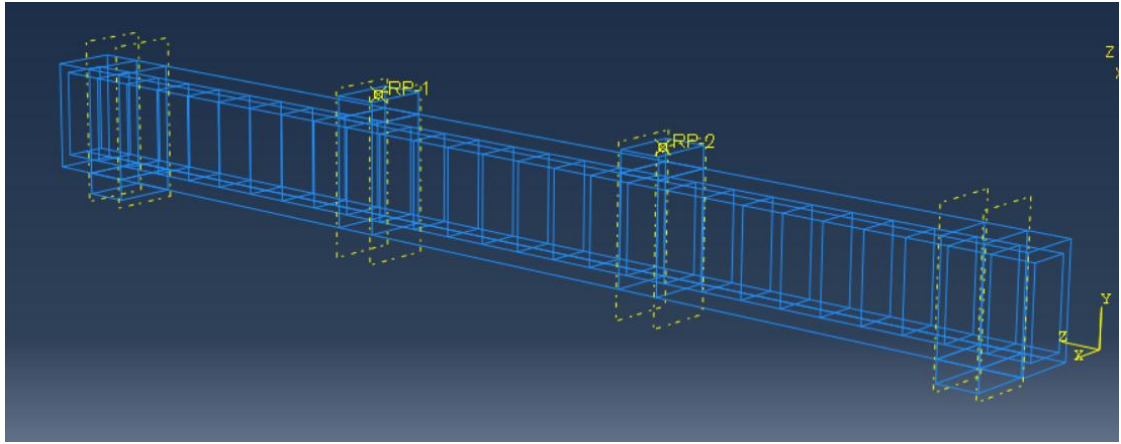

Figure 2.3 Wireframe view of the FE model highlighting reinforcement and reference points (RPs).

### 2.2.4 Material models

Concrete is modelled using the Concrete Damaged Plasticity model. The uniaxial compressive and tensile constitutive curves and the associated damage evolution are defined using the calibration procedure proposed by Li and Zhang [35]. Reinforcing steel is modelled as an isotropic elastic plastic material. The elastic modulus and yield properties are assigned based on the reference specimen data described in Section 2.1.

### 2.2.5 Boundary conditions and loading control

Simply supported boundary conditions are applied at two support regions to achieve a clear span of 2400 mm (Figure 2.4). To prevent rigid body motion, one support constrains the vertical displacement, while the other support additionally constrains one longitudinal degree of freedom.

The load is applied through two loading blocks. Displacement control is used by prescribing the vertical displacement at the reference points of the loading blocks. Reaction forces are collected as RF2 at the loading block reference points. The total applied load is obtained by summing the RF2 values from the two loading points. For the parametric simulation campaign, each loading block is allowed to relocate along the beam span by up to 200 mm to the left or right relative to its baseline position.

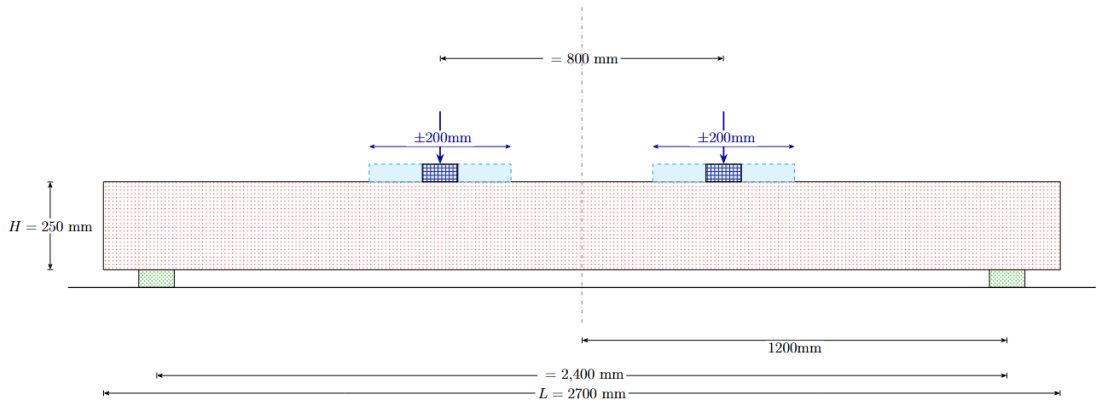

Figure 2.4 Boundary conditions and four-point bending setup of the RC beam model.

2.2.6 Analysis step and exported outputs

A quasi-static nonlinear analysis step Static General is used with geometric nonlinearity. Automatic incrementation is adopted to improve convergence during stiffness degradation. Field outputs are saved for each recorded increment, including nodal displacement U, element wise von Mises stress and element-wise equivalent plastic strain. These outputs, together with the reaction force history from the loading reference points, are exported to a unified data format for subsequent surrogate training and evaluation.

## 2.3 Parametric FE dataset generation and exported responses

To construct a dataset for training and evaluating the GNN surrogate, we performed a parametric simulation campaign based on the baseline Abaqus model described in Section 2.2. In total, 190 FE cases were generated by relocating the two loading blocks in the four-point bending setup. The two loading blocks were allowed to move independently, so the loading configuration was not constrained to be symmetric. For each loading block, the maximum relocation range was limited to 200 mm relative to its baseline position.

The concrete model uses a 25 mm global mesh size. To ensure that load application remains aligned with the discretised mesh and to simplify automated post-processing, the loading block locations were restricted to integer multiples of 25 mm along the beam span.

All cases were simulated under displacement control by prescribing the vertical displacement at the reference points of the two loading blocks. This loading strategy was adopted to ensure stable convergence and to obtain a complete nonlinear response history for each loading configuration. The global load history was computed as the sum of the vertical reaction forces (RF2) extracted at the two loading reference points at each sampled frame.

Abaqus analyses were performed with automatic time incrementation, yielding approximately 750 increments per case. To store a compact yet representative response trajectory for surrogate learning, we did not export outputs at every increment. Instead, we sampled the response at fixed normalised loading progress levels $p \in \{0, 0.05, 0.10, ..., 1.00\}$, resulting in 21 frames per case (including the initial frame at $p = 0$). Here $p$ denotes the normalised imposed displacement level, defined as the ratio between the current prescribed displacement and the maximum prescribed displacement of that case. The sampled frames are aligned and indexed consistently across all exported quantities.

At each sampled frame, we exported: (i) nodal displacement $U(p)$ (three displacement components for all nodes), (ii) element-wise von Mises stress $S_{\text{Mises}}(p)$ for the C3D8R concrete mesh, (iii) element-wise equivalent plastic strain $\text{PEEQ}(p)$, and (iv) the global reaction force history $RF2(p)$ (sum of the two loading reference points), as listed in Table 2.

Table 2 Summary of the FE dataset and experimental setup

| Item | Value |
| --- | --- |
| frames per case | 21 |
| Element type (concrete) | C3D8R |
| Global mesh size | 25 mm |
| Loading positions | 2 loading blocks, mesh-aligned (25 mm grid) |
| Outputs | $U(t)$ (node), $s_e(t)$ (elem), $\text{PEEQ}_e(t)$ (elem), $RF2(t)$ (global) |
| RF2 definition | sum of RF2 at two loading reference points |
| Split protocol | case-level split |

## 3. Graph Neural Network Surrogate Method

### 3.1 Problem formulation and model overview

This section proposes a dual-graph spatiotemporal graph neural network (GNN) surrogate to approximate the time-evolving finite element (FE) responses of a concrete beam under four-point bending with varying loading-block locations. Based on the dataset described in Section 2 (stored in NPZ files containing mesh information, full-field histories, and global signals), the surrogate performs autoregressive rollout over time and jointly predicts (i) nodal displacement fields, (ii) element-wise von Mises stress fields, and (iii) the global vertical reaction force history.

### 3.1.1 Task definition and notation

For each FE case, the mesh consists of $N$ nodes and $E$ C3D8 elements, and the response is sampled at $T$ frames. At each frame $t$, the surrogate is trained to predict four quantities: the nodal displacement field $\mathbf{u}_t \in \mathbb{R}^{N\times 3}$, the element-wise von Mises stress $\mathbf{s}_t \in \mathbb{R}^E$, the element-wise equivalent plastic strain (PEEQ) $\mathbf{p}_t \in \mathbb{R}^E$, and the global vertical reaction force $\text{RF2}_t \in \mathbb{R}$. Here $\text{RF2}_t$ is obtained by summing the vertical reaction forces at the reference points of the two loading blocks at frame $t$. The model conditions these predictions on (i) the mesh description (nodal coordinates and element connectivity), (ii) an encoding of the loading configuration (e.g., indicators for the two load regions), and (iii) a normalized loading progress variable $\alpha_t \in [0,1]$ that indexes the sampled frames.

### 3.1.2 Overall pipeline

The proposed method follows four main stages:

1) Dual-graph representation. The FE mesh is represented by a node graph and an element graph. The node graph supports displacement modelling at the nodal resolution, while the element graph supports local interactions for stress evolution and provides a basis for global response regression.

2) Spatiotemporal recurrent modelling. Two recurrent graph modules (GConvGRU) evolve hidden states on the node and element graphs across time and enable rollout.

3) Multi-task outputs. The node branch predicts $\hat{\mathbf{u}}_t$. The element branch predicts $\hat{\mathbf{s}}_t$ and regresses the scalar $\widehat{RF2}_t$ from pooled element representations.

4) Multi-task training. Joint supervision on $\mathbf{u}_t$, $\mathbf{s}_t$, $\text{peeq}_t$ and $RF2_t$ is used to optimize a shared surrogate that matches both full-field and global FE responses.

Implementation details on graph construction, feature definition, normalisation, and batching are provided in Section 3.2 to ensure reproducibility.

### 3.1.3 DualGraph-GConvGRU surrogate architecture

The surrogate adopts a coupled node branch and element branch architecture. Both recurrent modules use a hidden size of 256, and the node-graph message passing uses $K = 2$ hops in the GConvGRU operator, consistent with the training configuration.

(1) Node branch: displacement prediction.

On the node graph $\mathcal{G}_n$, a GConvGRU updates nodal hidden states $H_t^n$ and an MLP decodes the displacement field:

$$\mathrm{H}_t^n = \mathrm{GConvGRU}_n(\mathrm{X}_t, \mathcal{G}_n; \mathrm{H}_{t-1}^n), \quad \hat{u}_t = \mathrm{MLP}_u(\mathrm{H}_t^n). \tag{1}$$

Here $\mathrm{X}_t$ is the node feature matrix at frame $t$, defined in Section 3.2.

(2) Element branch: stress and reaction-force prediction.

The element branch receives element inputs obtained by aggregating node hidden states onto elements, then evolves element hidden states $\mathrm{H}_t^e$ on the element graph $\mathcal{G}_e$ using another GConvGRU, and decodes element stress:

$$\mathrm{Z}_t^e = \mathrm{Agg}_{n \to e}(\mathrm{H}_t^n), \quad \mathrm{H}_t^e = \mathrm{GConvGRU}_e(\mathrm{Z}_t^e, \mathcal{G}_e; \mathrm{H}_{t-1}^e), \\ \hat{s}_t = \mathrm{MLP}_s(\mathrm{H}_t^e) \tag{2}$$

To predict the global reaction force, the element hidden states are pooled into a global representation and regressed by an MLP:

$$\widehat{RF2}_t = \mathrm{MLP}_{rf2}(\mathrm{Pool}(\mathrm{H}_t^e)). \tag{3}$$

This design aligns with the interpretation of $RF2$ as a global structural response scalar.

3.1.4 Autoregressive rollout

The model performs autoregressive rollout along time. After producing $\hat{u}_t$, $\hat{s}_t$, and $\widehat{RF2}_t$ at frame $t$, the next-frame node inputs are formed by feeding back the predicted state, and the recurrent updates proceed to frame $t+1$. With this mechanism, inference requires only the static conditions (mesh and loading-condition encoding) and the progress variable $\alpha_t$ to generate a complete response trajectory.

3.1.5 Multi-task objective

Training minimizes a joint objective over displacement, stress, and reaction force:

$$\mathcal{L} = \mathcal{L}_u + \lambda_s \mathcal{L}_s + \lambda_{rf2} \mathcal{L}_{rf2} + \lambda_p \mathcal{L}_p + \lambda_{lap} \mathcal{L}_{lap}, \tag{4}$$

where $\mathcal{L}_u$ is the displacement error (MSE), $\mathcal{L}_s$ is the stress error, $\mathcal{L}_p$ is the equivalent plastic strain error, and $\mathcal{L}_{rf2}$ is the reaction-force error (MSE). $\mathcal{L}_{lap}$ is an optional Laplacian smoothing regulariser applied to the predicted nodal displacement field over the node graph. The weights $\lambda_s$, $\lambda_{rf2}$, $\lambda_p$, and $\lambda_{lap}$ balance the multi-task terms.

## 3.2 Graph construction and feature definition

This section describes how each NPZ case is converted into (i) a node graph, (ii) an element graph, and (iii) time-indexed node feature tensors $X_t$. Each case stores the FE mesh (nodal coordinates and C3D8 connectivity) and time-sampled responses over $T$ frames, including nodal displacement, element-wise von Mises stress, element-wise PEEQ, and the global vertical reaction force. The NPZ file also provides the node set on the load-application surface and the frame times used to compute the normalized progress variable $\alpha_t$. These quantities are used to construct the graphs and $X_t$ as described below.

### 3.2.1 Graph construction

The FE mesh is represented using two graphs. The node graph $G_n = (V_n, E_n)$ takes FE nodes as vertices. Node adjacency is derived from the C3D8 element connectivity: within each hexahedral element, edges are added between the 12 standard corner-node pairs that define the element's edges. Collecting these connections over all elements yields the node-level neighbourhood used for message passing. The graph is treated as undirected by including both directions for each connected node pair.

The element graph $G_e = (V_e, E_e)$ takes C3D8 elements as vertices. Two elements are connected if they share a common face, i.e., the same set of four corner nodes. Face sharing is identified by comparing the node sets that define element faces across neighbouring elements, and the resulting adjacency is again treated as undirected. Face adjacency is adopted because it reflects direct local coupling between neighbouring elements for stress and internal-variable evolution, while avoiding overly dense connectivity that can arise when elements are connected merely through sharing a node. The C3D8 node numbering convention used to enumerate element edges and faces is shown in Figure 3.1.

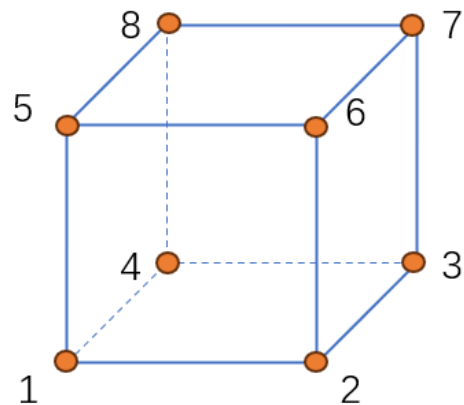

Figure 3.1 Node numbering convention for a C3D8 hexahedral element.

### 3.2.2 Cross-scale coupling

Element–node incidence is given by the FE connectivity, where each C3D8 element is associated with its eight corner nodes. This incidence relation is used to couple the node and element branches. Specifically, the element branch receives an element-level input obtained by mean pooling the node hidden states over the eight incident nodes of each element. This operation transfers kinematic information learned on the node graph to the element-level predictor in a mesh-consistent manner (Figure 3.1).

3.2.3 Feature construction and conditioning variables

To learn across cases with different loading-block locations, the surrogate is explicitly conditioned on the loading configuration and loading progress. The loading configuration is encoded by a binary node indicator that marks whether a node lies on the load-application surface; this indicator is included as a node feature at every frame so that the network can distinguish different load locations across cases. Loading progress is represented by a normalized variable $\alpha_t \in [0,1]$ computed from the recorded frame times by linearly mapping the first and last frame of each case to 0 and 1, respectively, and broadcasting $\alpha_t$ to all nodes at frame $t$.

At each frame $t$, the node feature matrix $X_t$ is formed by concatenating normalized nodal coordinates, the previous-step displacement state, $\alpha_t$, the load-region indicator, and a velocity-like increment computed from displacement differences between consecutive frames. Optionally, a nodal scalar derived from the previous-step element stress may be provided by averaging element values to their incident nodes; this optional feedback is used only as an input feature and does not replace element-level supervision. For the initial frame, history-dependent terms are initialized to zeros to start the rollout. Continuous quantities are normalized for training stability using either global statistics computed from the training set or per-case statistics, and the same normalisation is applied consistently during evaluation.

Mini-batch training combines multiple cases by concatenating their graphs with case-wise index offsets to avoid index collisions, producing a single merged graph for one forward pass while preserving per-case connectivity and incidence relations. An overview of the coupled node–element surrogate architecture is provided in Figure 3.2.

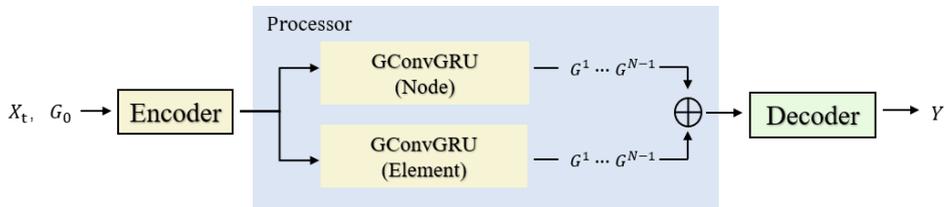

Figure 3.2 Architecture of the proposed Dual-Graph GConvGRU surrogate model.

## 3.3 Training strategy

This section presents the training protocol, including the autoregressive rollout and the multi-task objective in Eq. (4). Training hyperparameters are reported in Table 3.

### 3.3.1 Autoregressive rollout

The surrogate is trained to generate the full response trajectory in an autoregressive manner. Given the input features at time step $t$, the model predicts $\hat{\mathbf{u}}_t$, $\hat{\mathbf{s}}_t$, $\hat{\mathbf{p}}_t$, and $\widehat{RF2}_t$. The predicted states are then used to form the inputs for the next step, enabling multi-step rollout over all sampled frames. This training setup matches the inference mode, where the model must progressively propagate its own predicted states to produce a complete trajectory for unseen loading configurations.

### 3.3.2 Objective and optimisation

The overall training objective is defined in Eq. (4) as a weighted multi-task loss. All task losses are computed using mean squared error (MSE) at their corresponding resolutions (nodal displacement at nodes; stress and PEEQ at elements; reaction force as a scalar sequence). In addition, we include a Laplacian regularisation term on the predicted displacement field to encourage spatial smoothness:

$$\mathcal{L}_{lap} = \sum_{i=1}^{N} \| \mathbf{u}_i - \frac{1}{d_i} \sum_{j \in \mathcal{N}_i} \mathbf{u}_j \|_2^2 \qquad (5)$$

where $\mathcal{N}_i$ denotes the one-hop neighbourhood of node $i$ and $d_i = |\mathcal{N}_i|$.

We optimise the network parameters using Adam. A validation set is used for model selection, and the checkpoint with the lowest validation loss is retained for final testing. To stabilise training, gradient clipping is applied, and the learning rate is adaptively reduced based on the validation-loss plateau, as specified in Table 3.

Table 3 Hyperparameters and training configuration

| Setting | Value |
|---|---|
| Node hidden size | 256 |
| Element hidden size | 256 |
| Message passing hops (K) | 2 |
| Epochs | 1000 |
| Batch size (cases) | 8 |
| Optimizer | Adam |
| Learning rate | 3e-3 |
| LR scheduler | ReduceLROnPlateau (patience=3, factor=0.5) |
| Gradient clipping | 0.5 |
| Normalisation | global z-score |

## 3.4 Ablation baseline and fair comparison protocol

This section defines the node-only single-graph baseline and the fair comparison protocol used to isolate the benefit of element-level supervision enabled by the dual-graph design.

3.4.1 Motivation and baseline definition

A key design choice of the proposed surrogate is to supervise and predict stress-like internal variables directly at the element resolution. In our Abaqus dataset, von Mises stress and PEEQ are exported as element-wise fields on the C3D8R mesh. To isolate the benefit of element-level supervision, we construct a single-graph baseline that operates only on the node graph by removing the element branch. The baseline predicts nodal displacement $U(t)$ in the same way as the full model, but cannot be directly supervised on element-wise stress/PEEQ without introducing a projection between element and node resolutions.

3.4.2 Element to Node to Element supervision mapping

Full model (Dual-Graph). The full model (DualGraph-GConvGRU) jointly predicts: (i) nodal displacement $U(t)$, (ii) element-wise stress $s_e(t)$, (iii) element-wise PEEQ $p_e(t)$, and (iv) the global reaction force history $RF2(t)$. Element representations are formed by aggregating the hidden states of the eight corner nodes of each element:

$$H_e^{in}(t) = \frac{1}{8} \sum_{n \in \mathcal{V}(e)} H_n(t), \tag{6}$$

Element inputs are constructed by averaging the hidden states of the eight corner nodes associated with each element, yielding an element-level representation. Stress and PEEQ are decoded from this element representation, while $RF2(t)$ is predicted by applying a global pooling operator over all element representations followed by a regression head.

Single-graph baseline (node-only). The baseline predicts nodal proxies $\hat{s}_n(t)$ and $\hat{p}_n(t)$ decoded from node hidden states. To compute losses at the element resolution (so that supervision scale remains comparable), we apply an Element to Node to Element mapping:

$$s_n(t) = \frac{1}{|\mathcal{E}(n)|} \sum_{e \in \mathcal{E}(n)} s_e(t), \quad p_n(t) = \frac{1}{|\mathcal{E}(n)|} \sum_{e \in \mathcal{E}(n)} p_e(t) \tag{7}$$

where $\mathcal{E}(n)$ denotes the set of elements incident to node $n$.

- Node-to-element reconstruction (predictions):

$$\hat{s}_e(t) = \frac{1}{8} \sum_{n \in \mathcal{V}(e)} \hat{s}_n(t), \quad \hat{p}_e(t) = \frac{1}{8} \sum_{n \in \mathcal{V}(e)} \hat{p}_n(t). \tag{8}$$

The stress/PEEQ losses are then computed between $\{\hat{s}_e(t), \hat{p}_e(t)\}$ and the ground-truth $\{s_e(t), p_e(t)\}$. This ensures that the comparison differs only in the representational route (node-only with averaging vs. dual-graph with direct element supervision), rather than a change in supervision resolution.

3.4.3 Fair comparison protocol

Unless explicitly stated otherwise, the dual-graph model and the single-graph baseline share identical settings, including: (i) dataset split and global z-score normalisation, (ii) node input features, (iii) recurrent operator family and comparable capacity (hidden sizes and message-passing hops), and (iv) the same optimisation protocol and training budget. The concrete settings used for all comparisons are summarised in Table 3.

## 4. Experimental Result

### 4.1 Setup and evaluation protocol

All experiments follow the parametric FE dataset described in Section 2. The concrete domain is discretised with C3D8R elements (25 mm global mesh size). We generate 190 cases by varying the two loading-block locations in the four-point bending setup. To obtain compact and temporally aligned trajectories, FE outputs are sampled at 21 fixed normalised loading-progress levels (including the initial frame). For each frame, we export nodal displacement $U(t)$, element-wise von Mises stress $s_e(t)$, element-wise PEEQ $p_e(t)$, and the global reaction-force history $RF2(t)$ (defined as the sum of vertical reaction forces at the two loading reference points).

4.1.1 Case-level split and model selection

All data splits are performed at the case level to avoid temporal leakage. We randomly shuffle cases and construct training/validation/test subsets following the split protocol in Section 2. The best checkpoint is selected by the lowest validation objective during training and is used for final testing.

4.1.2 Normalisation and rollout setting

Continuous quantities are standardised using global z-score statistics computed on the training set only and then applied to validation and test sets. RF2 is normalised as an independent scalar target to prevent scale dominance in the multi-task objective. The model is evaluated under an autoregressive rollout over the 21 sampled frames, conditioned on a normalised progress variable $\alpha(t)$. Node inputs concatenate normalised geometry and history-dependent state features (e.g., previous displacement and displacement increments), together with progress conditioning and load-region encoding. When used, the stress-history feedback is constructed by distributing element stresses to incident nodes via averaging.

4.1.3 Metrics and aggregation

Performance is reported for three output types: nodal displacement $U(t)$, element-wise fields $s_e(t)$ and $p_e(t)$, and the global history $RF2(t)$. We compute MSE with mean reduction over all relevant dimensions (frames, spatial points, and components where applicable) and report RMSE as $RMSE = \sqrt{MSE}$. $R^2$ is additionally reported for full-field outputs. Unless otherwise stated, metrics are computed in the normalised space consistent with the optimisation objective; values in physical units can be recovered by reversing the z-score transformation. Training settings and hyperparameters are summarised in Table 3.

## 4.2 Main predictive performance of the proposed Dual-Graph model

Using the experimental protocol defined in Section 4.1, we evaluate the proposed Dual-Graph surrogate on the held-out test set. This section reports the main results of the full model; ablation against the single-graph baseline is presented in Section 4.3.

4.2.1 Overall accuracy on the test set

Table 4 summarises the quantitative performance of the proposed Dual-Graph model on the held-out test set. RMSE values are reported in physical units by reversing the z-score normalisation, and $R^2$ is computed between predicted and FE references using the same aggregation protocol as in Section 4.1. The model achieves low error for nodal displacement while maintaining strong agreement for element-wise von Mises stress and PEEQ. The global reaction-force history $RF2(t)$ is also predicted accurately, indicating that the learned field evolution is consistent with the overall structural response.

Table 4 Quantitative performance of the proposed Dual-Graph model on the test set

| Output | RMSE | Unit | R² |
| --- | --- | --- | --- |
| Displacement (U) | 0.324 | mm | 0.9969 |
| von Mises stress | 1.381 | MPa | 0.924 |
| PEEQ | 1.56e-4 | — | 0.8915 |
| RF2 | 1.583 | kN | 0.9907 |

4.2.2 Global force–deflection agreement

Figure 4.1 compares the predicted and FE force–deflection responses for representative test cases, where $RF2(t)$ is plotted against the midspan deflection $\delta_{\text{mid}}(t)$. The midspan deflection is measured as the vertical displacement of the node at the geometric centre of the span. The predicted curves closely follow the FE references across the full loading path, capturing both the initial near-linear regime and the subsequent nonlinear evolution. This agreement suggests that the surrogate preserves the global load–deformation relationship during autoregressive rollout, rather than only matching individual field snapshots.

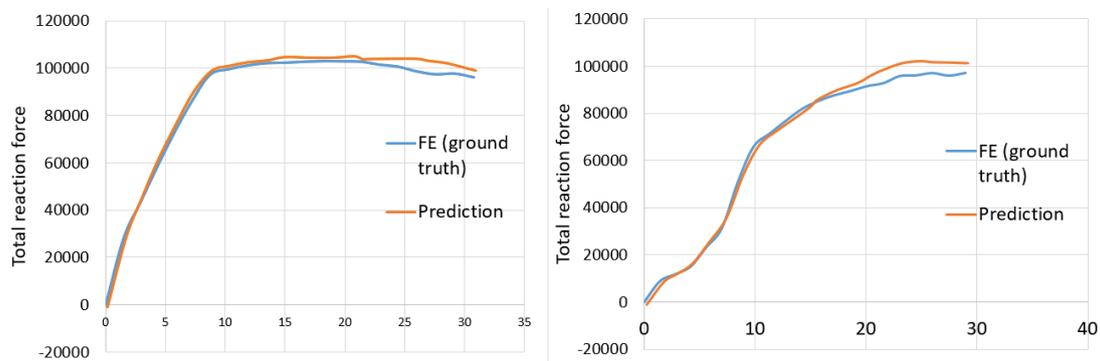

Figure 4.1 Predicted vs. FE Force-Deflection curves for representative test cases.

4.2.3 Spatial fidelity of element-wise stress and plastic strain fields

Figure 4.2 visualises the element-wise field predictions for a representative test case at selected loading-progress frames (peak). The proposed model reproduces the spatial distributions of von Mises stress and equivalent plastic strain (PEEQ), accurately capturing the locations and extents of high-stress bands and localised plasticity concentrations. The remaining discrepancies are primarily confined to high-gradient regions, where minor differences in peak magnitude or front sharpness can lead to larger pointwise errors, while the overall spatial topology of the fields is preserved.

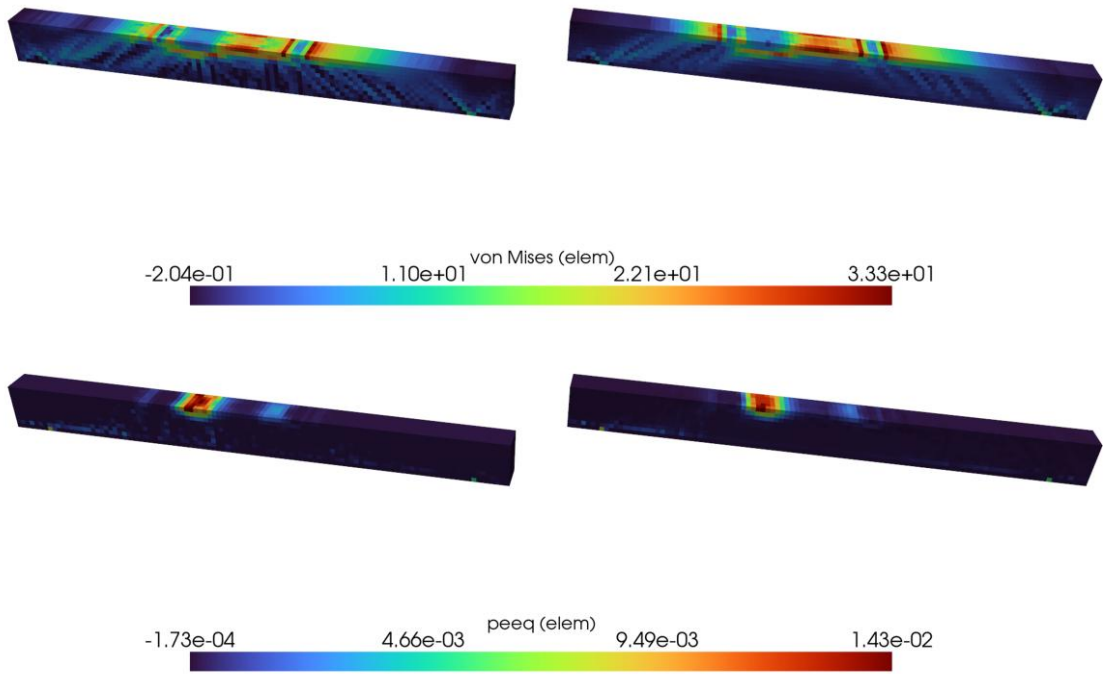

Figure 4.2 Von Mises stress and PEEQ fields (GT / Pred) at selected frames (case 119).

## 4.3 Ablation study: Dual-Graph vs Single-Graph baseline

4.3.1 Comparison setup and fairness

This section conducts a controlled ablation by comparing the proposed Dual-Graph model with a node-only Single-Graph baseline. Both models are trained and evaluated under identical settings, including the case-level split, normalisation strategy, training budget, and optimisation configuration. The only difference lies in the supervision granularity and representational route: the Dual-Graph model predicts stress-like internal variables directly at the element resolution, whereas the Single-Graph baseline learns node-level proxies and requires an Element to Node to Element conversion to form element-wise fields for supervision and evaluation. Consequently, this ablation isolates the effect of direct element-level modelling and supervision.

4.3.2 Peak attenuation induced by Element to Node to Element projection

To quantify the intrinsic information loss introduced by resolution conversion, we apply the Element to Node to Element mapping directly to the FE element-wise ground-truth fields of von Mises stress and PEEQ, without involving any trained model. This controlled procedure isolates the effect of spatial averaging alone.

Figure 4.3 compares the original element-level fields, the projected fields after Element to Node to Element mapping, and their differences for a representative case at a near-peak response frame. The peak von Mises stress decreases from 29.49 to 23.55 (a reduction of approximately 20.1%). Similarly, the peak PEEQ decreases from 0.0979 to 0.0780 (approximately 20.3%). The differences concentrate in localised high-gradient regions, indicating that the projection acts as a smoothing operator that systematically attenuates peak magnitudes. These results confirm that peak suppression is an inherent consequence of the Element to Node to Element conversion, rather than a training artefact.

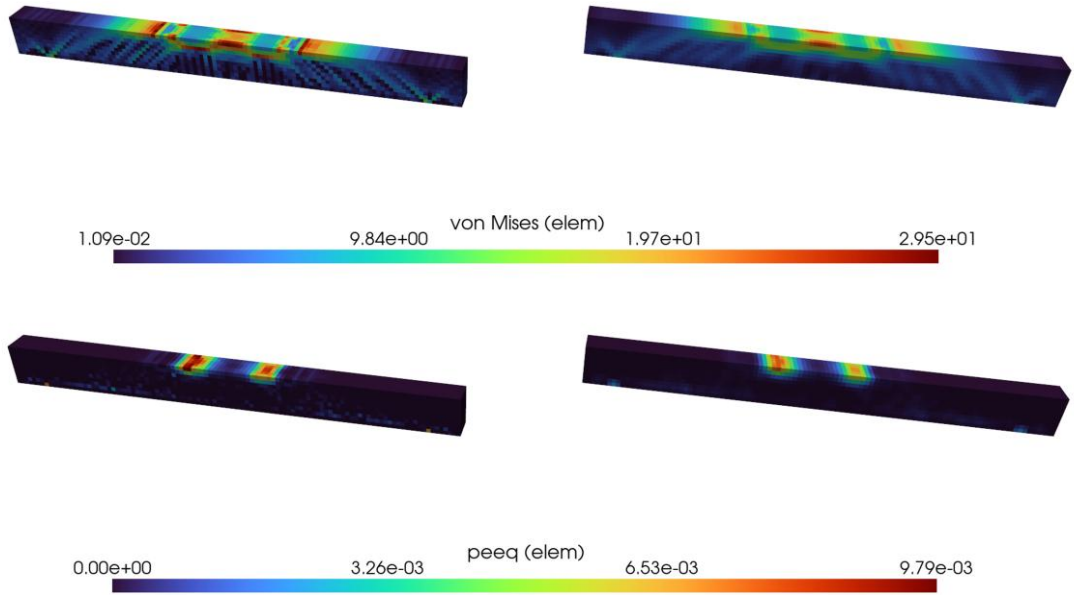

Figure 4.3 Effect of the Element to Node to Element projection on element-wise FE fields.

### 4.3.3 Prediction comparison: Single-Graph vs Dual-Graph

Having quantified the intrinsic peak attenuation introduced by the Element to Node to Element projection, we next compare the predictive performance of the Dual-Graph model and the node-only Single-Graph baseline under the same training protocol and model capacity. Quantitative results on the held-out test set are summarised in Table 5.

The Dual-Graph model consistently achieves lower errors for element-wise internal variables. Specifically, the element-wise von Mises stress RMSE decreases from 1.8602 to 1.3817, corresponding to a 25.7% reduction, and the PEEQ RMSE decreases from $2.1863 \times 10^{-4}$ to $1.5627 \times 10^{-4}$, corresponding to a 28.5% reduction. These gains are obtained under identical data split, normalisation, and optimisation settings, suggesting that the performance difference primarily arises from the representational route: the Dual-Graph preserves element-level states and performs supervision directly at the element resolution, whereas the Single-Graph baseline relies on node-level proxies coupled with node–element averaging, which inherently suppresses local peaks. The substantial error reduction indicate that direct element-level modelling is important for learning highly localised nonlinear responses in stress and plastic strain.

Table 5 Quantitative comparison between Single-Graph and Dual-Graph models on the test set

| Model | Stress RMSE | PEEQ RMSE |
| --- | --- | --- |
| Single-Graph | 1.8602 | 2.1863e-4 |
| Dual-Graph | 1.3817 | 1.5627e-4 |
| Relative reduction (%) | 25.7% | 28.5% |

4.3.4 Qualitative comparison: Single-Graph vs Dual-Graph

Figure 4.4 presents side-by-side visualizations of the predicted von Mises stress and PEEQ fields for a representative test case at selected frames, comparing the Single-Graph baseline with the proposed Dual-Graph model.

For a fair comparison at the element resolution, the Single-Graph predictions are first produced at nodes and then reconstructed back to elements using the same node-to-element averaging scheme adopted in the Element to Node to Element projection. In contrast, the Dual-Graph model outputs element-wise fields directly.

Across frames, the Single-Graph baseline exhibits a characteristic "over-smoothing" pattern in high-gradient regions: localized concentrations become spatially diffused, peak zones appear less distinct, and sharp transitions around critical regions are weakened. The corresponding absolute error maps show that the dominant errors concentrate around the same localized zones where the ground-truth fields display steep gradients, rather than being uniformly distributed.

By comparison, the Dual-Graph model better preserves both the spatial localization and the contrast of the concentrated regions. The predicted high-response zones are more consistent with the FE ground truth in terms of (i) the spatial footprint of the concentrated region, (ii) the sharpness of local gradients, and (iii) the relative intensity between peak and surrounding areas. This visual evidence supports the quantitative finding that retaining an explicit element-level representation is advantageous for learning peak-sensitive, highly localized nonlinear internal variables.

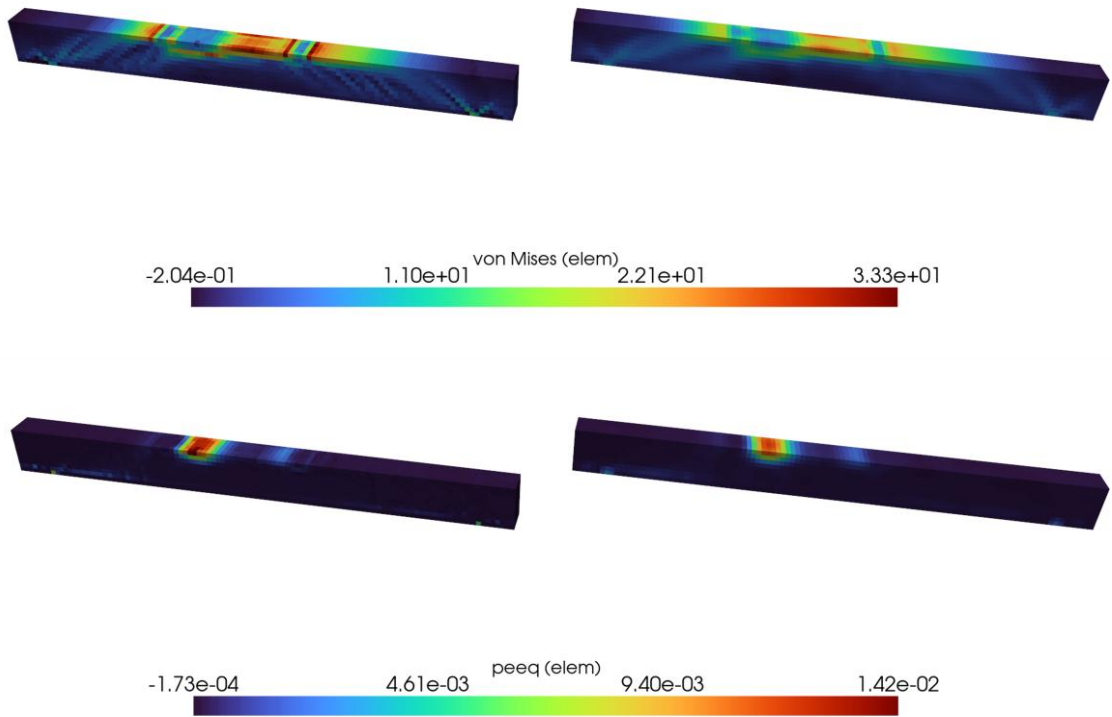

Figure 4.4 Comparison of predicted von Mises stress and equivalent plastic strain (PEEQ) fields between the Single-Graph and Dual-Graph models.

## 4.4 Spatial generalization of displacement fields across diverse loading configurations

4.4.1 Displacement field visualization across diverse loading cases

To examine robustness under varying loading configurations, Figure 4.5 compares the predicted and FE ground-truth displacement fields for multiple representative test cases. Each case corresponds to a distinct pair of loading-block positions (including both symmetric and asymmetric configurations), thereby covering diverse shear spans and constant-moment regions induced by the relocation of the two loading points. For each case, the global reaction force is computed as the sum of RF2 extracted at the two loading-block reference points, and the deflection is measured at the midspan, consistent with the evaluation protocol used throughout this section.

Across the representative cases, the proposed model reproduces the overall curve shape and key response characteristics, including the initial stiffness, the transition into nonlinear response, and the peak-load level. Minor discrepancies are primarily observed near highly nonlinear stages, where the global response becomes sensitive to localized damage/plasticity evolution. Nevertheless, the consistent agreement across diverse configurations indicates that the learned surrogate does not rely on a narrow subset of loading patterns and remains robust when loading positions vary within the designed parametric space.

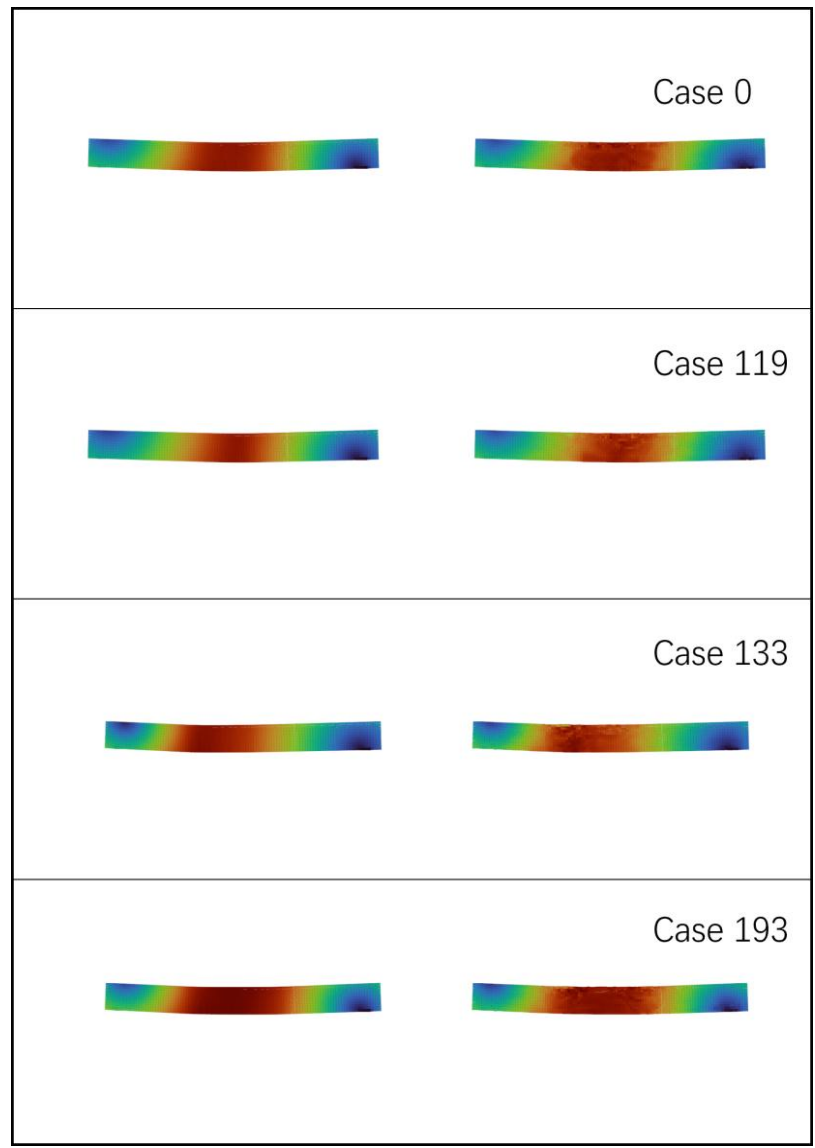

Figure 4.5 Qualitative comparison of displacement fields for multiple representative test cases with different loading-block positions.

## 5. Discussion and conclusions

(1) Main findings and efficiency.

This paper proposed a dual-graph spatiotemporal GNN surrogate (Dual-Graph GConvGRU) that explicitly couples a node graph and an element graph to jointly predict displacement, element-wise internal variables (von Mises stress and PEEQ), and the global reaction-force history for reinforced-concrete beams under four-point bending. The surrogate achieves markedly improved accuracy for peak-sensitive internal variables compared with a node-only baseline, while providing substantial computational acceleration. In particular, once trained, the model delivers predictions at inference time that are approximately two orders of magnitude faster (~100×) than running a full nonlinear FE analysis for the same case, enabling rapid parametric evaluation and design iteration.

(2) Why the dual-graph design improves internal-variable prediction.

The observed gains are consistent with the representational bottleneck introduced when element-wise targets must be learned through node-only representations. Element to Node to Element conversion acts as a spatial averaging operator that systematically smooths high-gradient regions and attenuates peaks, which are precisely the regions that dominate stress/PEEQ errors in nonlinear RC responses. By maintaining a dedicated element graph and applying direct element-level supervision, the proposed model avoids this unavoidable smoothing bottleneck and better preserves localization intensity, spatial footprint, and sharp gradients of internal-variable fields.

(3) Practical implications for mesh-based surrogates.

The results suggest a clear modelling guideline: node-only surrogates are often sufficient for smooth kinematic fields (e.g., displacement), whereas explicit element-level representations become critical when the targets are localized, history-dependent internal variables that govern damage/plasticity evolution. From an engineering workflow perspective, the combination of improved peak-sensitive accuracy and ~100× inference speedup makes the surrogate suitable for large parametric sweeps, sensitivity studies, and near-real-time "what-if" analyses where repeated FE runs would be prohibitive.

(4) Limitations and future work.

This study focuses on a single RC beam configuration and a bounded parametric variation achieved by relocating the two loading blocks along the span. Generalization beyond the current geometry, reinforcement detailing, mesh density, material parameters, and loading families remains to be systematically validated. Future work will extend the dataset coverage to broader structural variations, incorporate uncertainty quantification for robust decision support, and explore stronger physics guidance (e.g., equilibrium-consistency or constraint-aware losses) to further improve stability around peak and post-peak regimes, where responses are most sensitive to localized nonlinear evolution.